\documentclass[runningheads]{llncs}

\usepackage[utf8]{inputenc} 
\usepackage[T1]{fontenc}    
\usepackage{tikz}
\usepackage{lmodern}
\usepackage[breaklinks=true,bookmarks=false, colorlinks, linkcolor=red]{hyperref}
\usepackage{url}            
\usepackage{booktabs}       
\usepackage{amsfonts}       
\usepackage{nicefrac}       
\usepackage{microtype}      
\usepackage{marvosym}

\usepackage{graphicx}
\usepackage{lipsum}
\usepackage{authblk}

\usepackage{amsmath}
\usepackage{amssymb}
\usepackage{bm}
\usepackage{todonotes}
\usepackage{multirow}
\usepackage[ruled,vlined]{algorithm2e}

\usepackage{subfigure}

\usepackage{booktabs}
\usepackage{xspace}
\usepackage{cleveref}
\usepackage{pifont}
\usepackage{floatrow}
\floatstyle{plaintop}
\restylefloat{table}

\makeatletter
\DeclareRobustCommand\onedot{\futurelet\@let@token\@onedot}
\def\@onedot{\ifx\@let@token.\else.\null\fi\xspace}
\def\eg{\emph{e.g}\onedot} 
\def\ie{\emph{i.e}\onedot}

\def\etal{\emph{et al}\onedot}
\renewcommand{\paragraph}{%
	\@startsection{paragraph}{4}{\z@}%
	{0.1em \@plus 0.5ex \@minus 0.2ex}{-1em}%
	{\normalsize\bf}%
}
\makeatother

\newcommand{\xmark}{\ding{55}}%

\newcommand{\bx}{\bm{x}}
\newcommand{\by}{\bm{y}}

\begin{document}

\pagestyle{headings}
\mainmatter
\def\ECCVSubNumber{11}  

\title{Knowledge Distillation for Multi-task Learning} 


\titlerunning{Knowledge Distillation for Multi-task Learning}
%

\author{Wei-Hong Li \and Hakan Bilen}
\authorrunning{Wei-Hong Li et al.}
\institute{\small VICO Group, University of Edinburgh, United Kingdom \\
\email{\{w.h.li, hbilen\}@ed.ac.uk}
}

\maketitle

\begin{abstract}

Multi-task learning (MTL) is to learn one single model that performs multiple tasks for achieving good performance on all tasks and lower cost on computation. Learning such a model requires to jointly optimize losses of a set of tasks with different difficulty levels, magnitudes, and characteristics (\eg cross-entropy, Euclidean loss), leading to the imbalance problem in multi-task learning. To address the imbalance problem, we propose a knowledge distillation based method in this work. We first learn a task-specific model for each task. We then learn the multi-task model for minimizing task-specific loss and for producing the same feature with task-specific models. As the task-specific network encodes different features, we introduce small task-specific adaptors to project multi-task features to the task-specific features. In this way, the adaptors align the task-specific feature and the multi-task feature, which enables a balanced parameter sharing across tasks. 
Extensive experimental results demonstrate that our method can optimize a multi-task learning model in a more balanced way and achieve better overall performance.
\end{abstract}

\section{Introduction}
\label{sec:intro}

The objective of multi-task learning (MTL)~\cite{caruana1997multitask,ruder2017overview} is to develop methods that can tackle a large variety of tasks within a single model.
MTL has multiple practical benefits. 
First, learning shared parameters across multiple tasks leads to representations that can be more data-efficient to train and also generalize better to unseen data.
Second, sharing parameters and computations across tasks can significantly reduce both training and inference time over running multiple individual models, which is especially important in platforms with limited computational resources such as mobile devices.
Therefore there is a growing interest in developing MTL methods and MTL has been successfully applied to machine learning problems in several fields including natural language processing~\cite{clark2019bam}, computer vision~\cite{bilen2016integrated,kokkinos2017ubernet} and speech recognition~\cite{seltzer2013multi}. 

There are at least two challenges to achieve better performance and efficiency with MTL.
The first one is to design a multi-task deep neural network architecture that shares only the relevant parameters across the tasks and keeps the remaining ones task-specific.
This is in contrast to the standard MTL methods that share all the layers except the last few ones across all the tasks.
This heuristic is possibly suboptimal when the tasks have different characteristics and goals (\eg semantically low and high-level tasks), however, searching for an optimal architecture in an exponential configuration space is extremely expensive.
The second one is to develop MTL training algorithms that achieve good performance not only in one of the tasks but in all of them.
This problem is especially important when MTL involves jointly minimizing a set of loss functions for various problems with different difficulty levels, magnitudes, and characteristics (\eg cross-entropy, Euclidean loss). 
Thus a naive strategy of uniformly weighing multiple losses can lead to sub-optimal performances and searching for optimal weights in a continuous hyperparameter space can be prohibitively expensive. 

Concerned with the second problem, previous work \cite{chen2017gradnorm,sener2018multi,kendall2018multi,guo2018dynamic,liu2019end} addresses the \emph{unbalanced loss optimization} problem with balanced loss weighting and parameter updating strategies.
Kendall~\etal~\cite{kendall2018multi} weigh loss functions by considering the task-dependent uncertainty of the model at training time.
Sener~\etal~\cite{sener2018multi} pose the MTL as a multiple objective optimization problem and propose an approximate Pareto optimization method that uses Frank-Wolfe algorithm to solve the constrained optimization.
Yu~\etal~\cite{yu2020gradient} project the gradients for each loss function to a space where conflicting gradient components are removed to eliminate the disturbance between the tasks.
Although the previous work improves over the uniform weighing loss strategy in MTL, they still suffer from the problem of one task dominating the remaining ones and lower task performance than the single task models in standard multi-task benchmarks.

\begin{figure}[t]
	\centering
	\includegraphics[width=.99\linewidth]{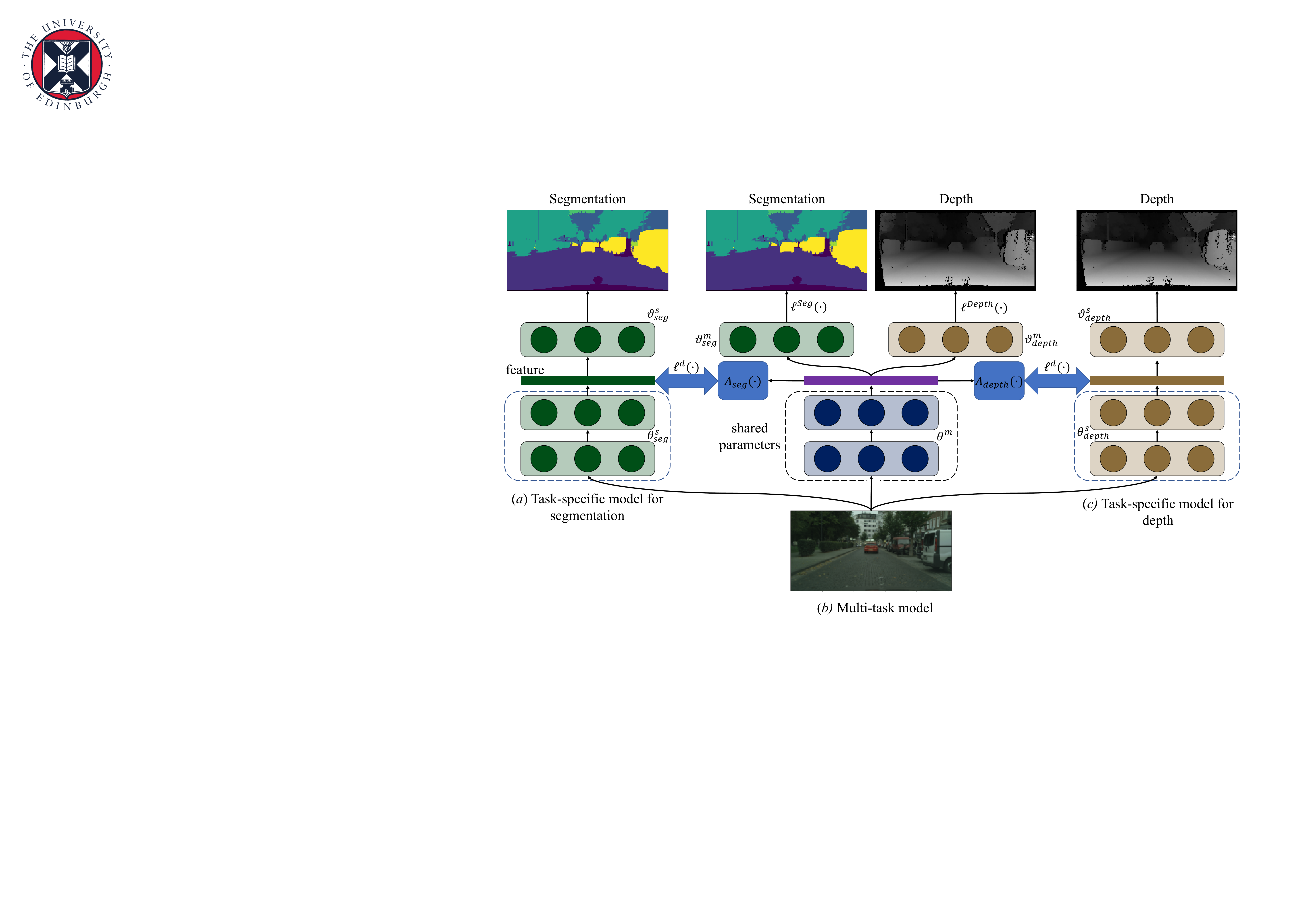}  
	\caption{Diagram of our method. We first train a task-specific model for each task in an offline stage and freeze their parameters (\ie (a), (c)). We then optimize the parameters of the multi-task network for minimizing a sum of task-specific losses and also for producing similar features with the single-task networks (\ie (b)). Best seen in color.}
	\label{fig:diagram}
	\vspace{-.1cm}
\end{figure}

In this paper, we approach the unbalanced MTL problem from a different point and propose a knowledge distillation based method inspired from~\cite{romero2014fitnets,hinton2015distilling}.
As weighing the individual loss functions (\eg \cite{kendall2018multi}) or modifying the gradients for the loss functions by simple transformations (\eg \cite{chen2017gradnorm,yu2020gradient}) provide a limited control on the learned parameters and are thus limited to prevent one task dominating the rest, we propose a more strict control on the parameters of the multi-task network.
Given that single-task networks often perform well with sufficient training data, we hypothesize that the solution of the multi-task network should be close to the single task ones' and lie in the intersection of the single-task solutions.
To this end, we first train a task-specific model for each task in an offline stage and freeze their parameters; then optimize the parameters of the multi-task network for minimizing a sum of task-specific losses and also for producing similar features with the single-task networks.
As each task-specific network can compute different features, we introduce small task-specific adaptors that map multi-task features to the task-specific one's.
The adaptors align the features of the single-task and multi-task networks, and enables a balanced parameter sharing across multiple tasks.

In the remainder of this paper, we first discuss how our method relates the previous MTL and data distillation methods in \Cref{sec:rel}, formulate our method in \Cref{sec:method}, demonstrate that our method outperforms the state-of-the-art MTL methods in two standard benchmarks in \Cref{sec:exp} and conclude the paper with future remarks in \Cref{sec:conc}.

\section{Related work}
\label{sec:rel}

\subsection{Multi-task Learning}

Multi-task learning (MTL) is one of the long-standing problems in machine learning and has been used broadly \cite{caruana1997multitask,rebuffi2017learning,ruder2017overview,du2018adapting,meyerson2018pseudo,meyerson2018pseudo,zhang2014facial,kendall2018multi,misra2016cross,kokkinos2017ubernet}. In computer vision, MTL has been used for image classification \cite{rebuffi2017learning}, facial landmark regression \cite{zhang2014facial}, segmentation and depth estimation \cite{kendall2018multi} and so on. In this work, we specifically focus on tackling the unbalance in the optimization of multi-task networks to achieve good performance not only in a few tasks but in all tasks.


In recent years, several methods have been proposed for solving the imbalance problem in MTL by either designing loss weighting schemes~\cite{chen2017gradnorm,kendall2018multi,sener2018multi,liu2019end,guo2018dynamic} to weigh each task-specific loss or modifying parameter updates~\cite{yu2020gradient}. Chen~\etal~\cite{chen2017gradnorm} develop a training strategy, namely GradNorm, that looks at the gradient's norm of each task and learns the weight to normalize each task's gradient so as to balance the losses for MTL. In \cite{sener2018multi}, Sener \etal formulate the MTL as a multiple objectives optimization problem and proposed an approximation Pareto optimization method using Frank-Wolfe algorithm to learn weights of losses.
Kendall~\etal~\cite{kendall2018multi} propose to weigh multiple loss functions by considering the homoscedastic uncertainty of each task during training. To design the weighting scheme, Guo \etal~\cite{guo2018dynamic} observe that the imbalances in task difficulty can lead to an unnecessary emphasis on easier tasks, thus neglecting and slowing progress on difficult tasks. Based on the observation, they introduce dynamic task prioritization for MTL, which allows the model to dynamically prioritize difficult tasks during training, where the difficulty is inversely proportional to performance. Rather than weighing the losses, Yu~\etal~\cite{yu2020gradient} propose a form of gradient ``surgery'' that projects each task's gradient onto the normal plane of the gradient of any other task and modifies the gradients for each task so as to minimize negative conflict with other task gradients during the MTL optimization. 

Unlike existing methods, we propose a knowledge distillation based MTL method to solve the unbalanced loss optimization problem from a different angle. To this end, we first train a task-specific model for each task in an offline stage and freeze their parameters. We then train the MTL network for minimizing task-specific loss and also for producing the same features with the task-specific networks. As the task-specific network encodes different features, we introduce small task-specific adaptors to project multi-task features to the task-specific features. In this way, the adaptors align the task-specific feature and the multi-task feature, which enables a balanced parameter sharing across tasks.

\subsection{Knowledge Distillation}

Our work is also related to knowledge distillation~\cite{hinton2015distilling,romero2014fitnets,tian2019contrastive,ma2019graph,phuong2019towards}.
Hinton~\etal~\cite{hinton2015distilling} show that distilling the knowledge of the whole ensemble of models to a neural network can achieve better performance and avoid an expensive computation. Romero~\etal~\cite{romero2014fitnets} introduce the knowledge distillation to training a small student network to achieve better performance than the teacher network. Apart from the success in single-task learning, knowledge distillation has also been shown to be effective in MTL. Parisotto~\etal~\cite{parisotto2015actor} exploits the use of deep reinforcement learning and model compression techniques to train a single policy network that learns to perform in multiple tasks by using the guidance of several expert teachers. In the contrast, in \cite{clark2019bam}, Clark \etal extends the Born-Again network \cite{furlanello2018born} to MTL setting for NLP. More specifically, they apply the knowledge distillation loss proposed in \cite{hinton2015distilling} on each task's predictions and propose a weight annealing strategy to update the weight of the distillation losses and multiple tasks losses.

Different from these methods, we aim at solving the unbalanced loss optimization problem in MTL. Aligning the predictions from the multi-task network and the task-specific networks would still result in unbalance as the dimension of tasks' predictions is usually different and we need to use different loss functions for matching different tasks' predictions \cite{clark2019bam}, \eg, a kl-divergence loss for classification and l2-norm loss for regression.
In this work, we first introduce a task-specific adaptor for each task to transform features from the multi-task network and we apply the same loss function to align the transformed multi-task feature and the task-specific networks' features. We train the MTL network for minimizing task-specific loss and for producing the same feature with task-specific networks. This enables the MTL to share the parameters in a balanced way.

\section{Methodology}
\label{sec:method}

\subsection{Single-task Learning (STL)}

Consider that we are given a dataset $\mathcal{D}$ that contains $N$ training images $\bx^i$ and their labels $\by_1^i,\dots,\by_T^i$ for $T$ tasks (\eg semantic segmentation, depth estimation, surface normals). 
In case of the STL, we wish to learn $T$ convolutional neural networks, one for each task, each maps the input $\bx$ to the target label $\by_\tau$, \ie $f(\bx;\theta^s_\tau,\vartheta^s_\tau)=\by_\tau$ where the superscript $s$ indicates the single-task, $\theta^s_\tau$ and $\vartheta^s_\tau$ are the parameters of the network.
Each single-task network is composed of two parts: i) a feature encoder $\phi(\cdot;\theta^s_\tau)$ that takes in an image and outputs a high-dimensional encoding $\phi(\bx;\theta^s_\tau)\in\mathbb{R}^{C\times H\times W}$ where $C$, $H$, $W$ are the number channels, height and width of the feature map; ii) a predictor $\psi(\cdot;\vartheta^s_\tau)$ that takes in the encoding $\phi(\bx;\theta^s_\tau)$ and predicts the output for the task $\tau$, \ie $\hat{\by}_\tau=\psi(\cdot;\vartheta^s_\tau)\circ\phi(\bx;\theta^s_\tau)$ where $\theta^s_\tau$ and $\vartheta^s_\tau$ denote the parameters of the feature encoder and predictor respectively.
The parameters for the network can be learned for each task independently by optimizing a task-specific loss function $\ell_\tau(\hat{\by},\by)$ (\eg Cross-Entropy loss function for classification) over the training samples that measure the mismatch between the ground-truth label and prediction as following:
\begin{equation}\label{eq:singletask}
	\min_{\theta^s_\tau,\vartheta^s_\tau}\sum_{\bx,\by_{\tau} \in \mathcal{D}}\ell_{\tau}(\psi(\cdot;\vartheta^s_\tau)\circ\phi(\bx;\theta^s_\tau),\by_{\tau}).
\end{equation}

\subsection{Multi-task Learning (MTL)}
In the case of MTL, we would like to learn one network that shares the majority of its parameters across the tasks and solves all the tasks simultaneously.
Similar to STL, the multi-task network can be decomposed into two parts: i) a feature encoder $\phi(\cdot;\theta^m)$ that encodes the input image into a high-dimensional encoding, now its parameters $\theta^m$ are shared across all the tasks; ii) a task-specific predictor $\psi(\cdot;\vartheta^m_\tau)$ for each task that takes in the shared encoding $\phi(\bx;\theta^m)$ and outputs its prediction for task $\tau$, \ie $\psi(\cdot;\vartheta^m_\tau)\circ\phi(\bx;\theta^m)$.
Note that we use superscript $m$ to denote MTL.
The multi-task network can be learned by optimizing a linear combination of task-specific losses:
\begin{equation}\label{eq:multitask}
	\min_{\theta^m,\vartheta^m_1,\dots,\vartheta^m_T}\sum_{\tau=1}^{T}\sum_{\bx,\by_{\tau} \in \mathcal{D}}w_{\tau}\ell_{\tau}(\psi(\cdot;\vartheta^m_\tau)\circ\phi(\bx;\theta^m),\by_{\tau})
\end{equation}
where $w_{\tau}$ is a scaling hyperparameter for task $\tau$ that is used for balancing the loss functions among the tasks.

In contrast to the STL optimization in \Cref{eq:singletask}, optimizing \Cref{eq:multitask} involves a joint learning of all the task-specific and shared parameters which is typically more challenging when the task-specific loss functions $\ell_{\tau}$ have different characteristics such as their magnitude and dynamics (\eg logarithmic, quadratic).
One solution to balance the loss terms is to search for the best scaling hyperparameters $w_\tau$ by a cross-validation which has two shortcomings.
First, the hyperparameter search in a continuous space is computationally expensive, especially when the number of tasks is large, as each validation step requires the training of the model.
Second, even when the optimal hyperparameters can be found, it may be sub-optimal to use the same fixed ones throughout the optimization.

\subsection{Knowledge Distillation for Multi-task Learning}
Motivated by these challenges, the previous work \cite{chen2017gradnorm,kendall2018multi,sener2018multi} propose dynamic weighing strategies that can adjust them at each training iteration.
Here we argue that these hyperparameters provide a limited control on the parameters of the network for preventing the unbalanced MTL and thus we propose a different view on this problem inspired by the knowledge distillation methods~\cite{romero2014fitnets,hinton2015distilling}.


To this end, we first train a task-specific model $f(\cdot;\theta^s_\tau,\vartheta^s_\tau)$ for each task $\tau$ by optimizing \Cref{eq:singletask} in an offline stage, freeze their parameters and use only their feature encoders $\phi(\cdot;\theta^s_\tau)$ to regulate the multi-task network at train time by minimizing the distance between the features of task-specific networks and multi-task network for given training samples (see \Cref{fig:diagram}). 
As the outputs of the task-specific encoders can differ significantly and the feature encoder of the multi-task network cannot match all of them simultaneously.
Instead, we project the output of the multi-task feature encoder into each task-specific one via a task-specific adaptor $A_{\tau}:\mathbb{R}^{C\times H\times W}\rightarrow\mathbb{R}^{C\times H\times W}$ where $H$, $W$ and $C$ are the height, width and depth (number of channels) of the features.
In our experiments, we use a linear layer that consists of a $1 \times 1 \times C\times C$ convolution for each adaptor.
These adaptors are jointly learned along the parameters of the multi-task network to align its features with the single-task feature encoders.

\begin{equation}\label{eq:distlossall}
	\mathcal{L}^d=\sum_{\tau=1}^{T}
	\sum_{\bx,\by_\tau \in \mathcal{D}}
			\ell^{d}(A_{\tau}(\phi(\bx;\theta^m)), \phi(\bx;\theta_{\tau}^s))
\end{equation} where $\ell^{d}$ is the Euclidean distance function between the L2 normalized feature maps:
\begin{equation}
	\ell^{d}(\bm{a},\bm{b})=\left\lVert\frac{\bm{a}}{||\bm{a}||_2}-\frac{\bm{b}}{||\bm{b}||_2}\right\lVert_2^2.
\end{equation}

Now we can write the optimization formulation that is employed to learn the multi-task model as a linear combination of  \Cref{eq:multitask} and \Cref{eq:distlossall}:
\begin{equation}\label{eq:KDMT}
	\min_{\theta^m,\vartheta^m_1,\dots,\vartheta^m_T}
	\sum_{\tau=1}^{T}\sum_{\bx,\by_{\tau} \in \mathcal{D}}w_{\tau}\ell_{\tau}(\psi(\cdot;\vartheta^m_\tau)\circ\phi(\bx;\theta^m),\by_{\tau}) + \lambda_\tau \ell^{d}(A_{\tau}(\phi(\bx;\theta^m)), \phi(\bx;\theta_{\tau}^s))
\end{equation} where $\lambda_\tau$ is the task-specific tradeoff hyperparameter.

\paragraph{Discussion.}
Alternatively, the inverse of each adaptor function can be thought as a mapping from each task-specific representation to a shared representation across all the tasks. 
The assumption here is that a large portion of encodings in the task-specific models is common to all the models up to a simple linear transformation.
While the assumption of linear relations between the features of highly non-linear networks may be surprising, such linear relations have also been observed in multi-domain \cite{rebuffi2018efficient} and multi-task problems \cite{stickland2019bert}.

\section{Experiments}
\label{sec:exp}

\subsection{Datasets}
We evaluate our method on three multi-task computer vision benchmarks, including SVHN \& Omniglot, NYU-v2, and Cityscapes.\footnote{The implementation of our method is available at \url{https://github.com/VICO-UoE/KD4MTL}.}

\vspace{0.1cm}
\paragraph{SVHN \& Omniglot} consists of two datasets, \ie SVHN~\cite{netzer2011reading} and Omniglot~\cite{lake2015human} where SVHN is a dataset for digital number classification and Omniglot is the one for characters classification. Specifically, SVHN contains 47,217 training images and 26,040 validation images of 10 classes. Omniglot consists of 19,476 training and 6492 validation samples of 1623 categories. As the testing labels are not provided, we evaluate all methods on the validation images and report the accuracy of both tasks. Note that in contrast to the NYU-V2 and Cityscapes datasets where each image is associated with multiple labels, each image in this benchmark is labeled only for one task. Thus the goal is to learn a multi-task network that can learn both tasks from SVHN and Omniglot.

\vspace{0.1cm}
\paragraph{NYU-V2 \cite{silberman2012indoor}} contains RGB-D indoor scene images, where we evaluate performances on 3 tasks, including 13-class semantic segmentation, depth estimation, and surface normals estimation. We use the true depth data recorded by the Microsoft Kinect and surface normals provided in \cite{eigen2015predicting} for depth estimation and surface normal estimation. All images are resized to $288 \times 384$ resolution as \cite{liu2019end}.

\vspace{0.1cm}
\paragraph{Cityscapes \cite{cordts2016cityscapes}} consists of street-view images, each labeled for two tasks: 7-class semantic segmentation\footnote{The original version of Cityscapes provides labels 19-class semantic segmentation. We follow the evaluation protocol in \cite{liu2019end}, we use labels of 7-class semantic segmentation. Please refer to \cite{liu2019end} for more details.} and depth estimation. We resize the images to $128 \times 256$ to speed up the training.

\subsection{Baselines}
In this work, we use the hard parameters sharing architecture for all methods where the early layers of the network are shared across all tasks and the last layers are task-specific (See \Cref{fig:diagram}). We compare our method with two baselines:
\begin{itemize}
	\item \textbf{STL} learns a task-specific model for each task.
	\item \textbf{Uniform}: This vanilla MTL model is trained by minimizing the uniformly weighted loss \Cref{eq:multitask}.
\end{itemize}

We also compare our method to the state-of-the-art MTL methods which are proposed for solving the unbalanced MTL, including Uncert~\cite{kendall2018multi}, MGDA~\cite{sener2018multi}, GradNorm~\cite{chen2017gradnorm} and a knowledge distillation based method, namely BAM~\cite{clark2019bam}, that applies knowledge distillation to network's prediction. On NYU-v2 and Cityscapes, we also compare our method with Gradient Surgery (GS)~\cite{yu2020gradient} and Dynamic Weight Average (DWA) \cite{liu2019end} with using different architectures, \ie SegNet \cite{badrinarayanan2017segnet} and MTAN~\cite{liu2019end} which is the extension of SegNet by introducing task-specific attention modules for each task.

\subsection{Comparison to the State-of-the-art}

\paragraph{Results on SVHN \& Omniglot.}

\begin{figure}
	\centering
	\includegraphics[width=.65\linewidth]{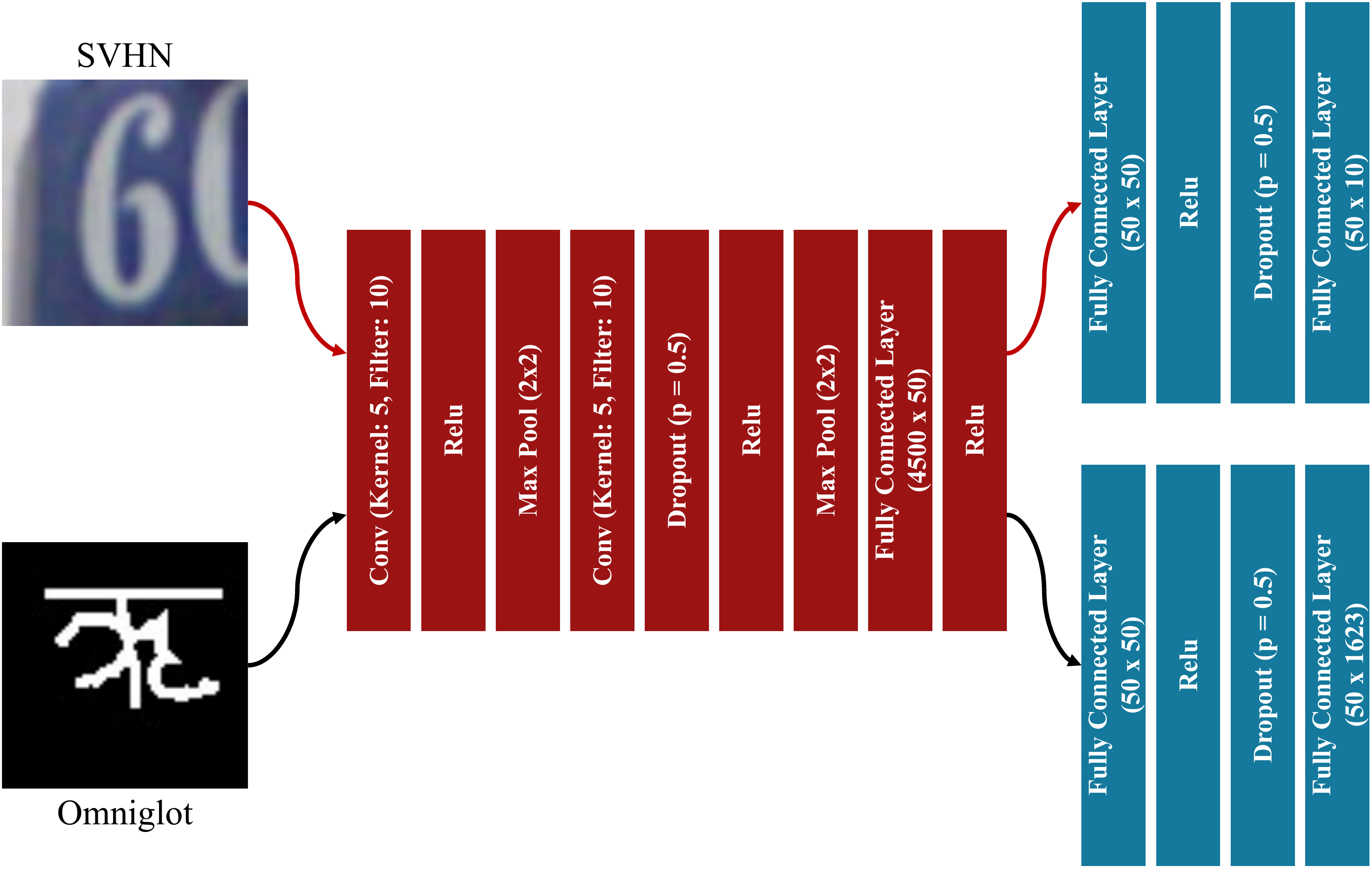}  
	\caption{Network architecture used in SVHN \& Omniglot.}
	\label{fig:LeNet}
	\vspace{-.1cm}
\end{figure}
First, we evaluate all methods on SVHN \& Omniglot. We extend the LeNet to MTL setting (See \Cref{fig:LeNet}) and use the extended network for all methods. We set the batch size of the mini-batch as 512 where 256 samples from SVHN and 256 images from Omniglot. We use Adam \cite{kingma2014adam} for optimizing the networks and adaptors. The learning rate of all task-specific adaptors is 0.01. We train all methods for 300 epochs in total where we scaled the learning rate by 0.85 every 15 epochs. In our method, weights of task-specific losses (\ie $w$ in \Cref{eq:KDMT}) are set uniformly. As a validation set for hyperparameter search ($\lambda$), we randomly pick 10\% of training data. After the best hyperparameters are chosen, we retrain with the full training set and report the median validation accuracy of the last 20 epochs in \Cref{tab:svhnomni}. We search over the set $\lambda=\{1,5,10,20\}$ of $\lambda$ and we chose $\lambda=10$.

\begin{table}[ht]
	\centering
	
    \resizebox{0.5\textwidth}{!}
	{
		\begin{tabular}{ccccc}


		\toprule
		Type & Methods & SVHN & Omniglot & avg \\
		\midrule
		\multirow{1}{*}{STL} & -\ - & 88.84 & 65.76 & -\ -\\
		\midrule
		\multirow{5}{*}{MTL} & Uniform & 85.88 & 66.91 & 76.40   \\
		& Uncert \cite{kendall2018multi} & 85.43 & 64.1 & 74.77 \\
		& MGDA \cite{sener2018multi} & 83.93 & 66.8 & 75.37 \\
		& GradNorm \cite{chen2017gradnorm} & 84.48 & 65.55 & 75.02 \\
		& BAM \cite{clark2019bam} & 86.57 & 66.08 & 76.33 \\
		& {\bf Ours} & {\bf 88.05} & {\bf 70.12} & {\bf 79.09}   \\
		\bottomrule
		\end{tabular}%
		\vspace{-0.1cm}
		\caption{Testing accuracy on SVHN \& Omniglot.}
		\label{tab:svhnomni}
	}
\end{table}%

\Cref{tab:svhnomni} shows that MTL with uniform loss weights (Uniform) obtains worse results on SVHN while it achieves better performance on Omniglot than STL. The state-of-the-art methods which dynamically weigh the task-specific losses cannot achieve a good trade-off between these two tasks. More specifically, Uncert and GradNorm obtain worse overall performance than STL while MGDA improves the performance on Omniglot and obtains worse performance on SVHN. Though BAM obtains better overall performance on both tasks, the improvement is achieved mainly because of more informative information provided by the continuous predictions of the teacher network (task-specific models). The unbalanced problem in MTL when we use BAM is still unsolved as BAM applies knowledge distillation on network predictions, which would have similar problems with the vanilla MTL (Uniform).
In contrast, our approach achieves significantly better performance than any other MTL methods \ie our method obtains 88.05 \% accuracy on SVHN and 70.12 \% accuracy on Omniglot. Compared with STL, our method obtains comparable results on SVHN and significant gains on Omniglot. These results strongly verify that our method is able to alleviate the unbalanced problem in this benchmark and to outperform STL as it enables the MTL model to learn more informative features.

\begin{table}[t]
	\centering
	
    \resizebox{1.0\textwidth}{!}
    {
		\begin{tabular}{cccccccccccc}


		    \toprule
		    \multirow{4}{*}{Architecture} & \multirow{4}{*}{Type} & \multirow{4}{*}{Methods} & \multicolumn{2}{c}{Segmentation} & \multicolumn{2}{c}{Depth} & \multicolumn{5}{c}{Surface Normal} \\
		    & & &  \multicolumn{2}{c}{\multirow{2}{*}{(Higher Better $\uparrow$)}}
		     & \multicolumn{2}{c}{\multirow{2}{*}{(Lower Better $\downarrow$)}}
		   & \multicolumn{2}{c}{Angle Distance} & \multicolumn{3}{c}{Within $t^{\circ}$}\\
		     & & & & & & &  \multicolumn{2}{c}{(Lower Better $\downarrow$)} & \multicolumn{3}{c}{(Higher Better $\uparrow$)}\\
		    & & & mIoU & Pix Acc & Abs Err & Rel Err & Mean & Median & 11.25 & 22.5 & 30 \\
		    \midrule
		    \multirow{6}{*}{SegNet} & STL & -\ - & 17.32 & 55.70 & 0.6577 & 0.2828 & 29.99 & 23.81 & 24.31 & 48.06 & 60.05 \\
		    \cmidrule{2-12}
		     & \multirow{5}{*}{MTL} & Uniform & 18.14 & 55.82 & 0.5841 & 0.2490 & 31.89 & 26.81 & 20.61 & 42.95 & 55.33 \\
		     & & Uncert \cite{kendall2018multi} & 16.79 & 52.51 & 0.6183 & 0.2612 & 32.44 & 27.27 & 20.81 & 42.41 & 54.47 \\
			 & & MGDA \cite{sener2018multi} & 15.57 & 49.05 & 0.6716 & 0.2710 & 29.90 & 24.13 & 23.81 & 47.58 & 59.75 \\
			 & & GradNorm \cite{chen2017gradnorm} & 18.08 & 55.76 & 0.5819 & 0.2495 & 30.65 & 25.38 & 22.64 & 45.34 & 57.59 \\
		     & & Ours & {\bf 18.75} & {\bf 58.02} & {\bf 0.5780} & {\bf 0.2467} & {\bf 29.40} & {\bf 23.71} & {\bf 24.33} & {\bf 48.22} & {\bf 60.45}\\
		 	\midrule
		 	
		 	\multirow{8}{*}{MTAN \cite{liu2019end}} & STL & -\ - & 16.38 & 53.89 & 0.6792 & 0.2963 & 30.66 & 24.26 & 23.35 & 47.34  & 59.33 \\
		 	\cmidrule{2-12}
		     & \multirow{7}{*}{MTL} & Uniform & 17.72 & 55.32 & 0.5906 & 0.2577 & 31.44 & 25.37 & 23.17 & 45.65 & 57.48 \\
		     & & DWA \cite{liu2019end} & 17.52 & 55.76 & 0.5869 & 0.2549 & 31.75 & 25.64 & 22.60 & 45.12 & 57.05   \\
		     & & Uncert \cite{kendall2018multi} & 17.67 & 55.61 & 0.5927 & 0.2592 & 31.25 & 25.57 & 22.99 & 45.83 & 57.67 \\
			 & & MGDA \cite{sener2018multi} & 15.60 & 52.36 & 0.6215 & 0.2767 & 30.26 & 24.01 & 23.98 & 47.77 & 59.85 \\
			 & & GradNorm \cite{chen2017gradnorm} & 17.37 & 55.92 & 0.5924 & 0.2630 & 31.20 & 24.91 & 23.11 & 46.27 & 58.20 \\
		     & & GS$^\ast$ \cite{yu2020gradient} & 20.17 & 56.65 & 0.5904 & 0.2467 & 30.01 & 24.83 & 22.28 & 46.12 & 58.77   \\
		     & & Ours & {\bf 20.75} & {\bf 57.90} & {\bf 0.5816} & {\bf 0.2445} & {\bf 29.97} & {\bf 23.96} & {\bf 24.24} & {\bf 47.78} & {\bf 59.78} \\
			\bottomrule
		\end{tabular}%
			}
		\vspace{-0.25cm}
		\caption{Testing results on NYU-v2. $^\ast$ Results of the `GS' method are from \cite{yu2020gradient}.}
		\label{tab:nyuv2}
\end{table}%

\paragraph{Results on NYU-V2.}
We follow the training and evaluation protocol in \cite{liu2019end}. We use cross-entropy loss for semantic segmentation, l1-norm loss for depth estimation, and cosine similarity loss for surface normal estimation. We train all methods using Adam \cite{kingma2014adam} with the learning rate initialized at 1e-4 and halved at the 100-th epoch for 200 epochs in total. The learning rate of all task-specific adaptors is 0.1 and the batch size is 2. In our method, weights of task-specific losses (\ie $w$ in \Cref{eq:KDMT}) are set uniformly. As a validation set for hyperparameter search ($\lambda$), we randomly pick 10\% of training data. After the best hyperparameters are chosen, we retrain with the full training set and report the results of three tasks on the validation set in \Cref{tab:nyuv2}. We search over the set $\lambda=\{1,2,3,4,5,6\}$ of $\lambda$ and the distillation loss' weight of Segmentation, depth, and surface normal are set to 1, 1 and 2, respectively.
From the results shown in \Cref{tab:nyuv2}, we can see that it is possible to tackle multiple tasks within a network and achieve performance improvement on some tasks, \eg when we use SegNet as the based network, the vanilla MTL (Uniform) achieves better performance on semantic segmentation and depth estimation though it causes a drop on surface normal estimation in comparison with STL. Though we see the benefits of using MTL, it is also clear that the unbalanced problem exists. 
We then apply existing methods that introduce loss weighting strategies for addressing the unbalanced loss optimization problem. From the results of using SegNet, GradNorm performs the best among all compared methods. However, it provides limited control on the learned parameters (\eg it achieves similar results to the MTL model using a uniformly weighting scheme) and still suffers from lower task performance than the single task models.

In comparison with these methods, our method obtains significant gains over all tasks and achieves better results than single task learning models. This strongly verifies our hypothesis that the solution of the multi-task network should be close to the single task ones' and lie in the intersection of the single-task solutions
and our method that applies stricter control on the parameters of the multi-task network can better address the unbalanced loss problem.


\begin{table}[t]
	\centering
	
    \resizebox{.7\textwidth}{!}
    {
		\begin{tabular}{ccccccc}


		    \toprule
		    \multirow{3}{*}{Architecture} & \multirow{3}{*}{Type} & \multirow{3}{*}{Methods} & \multicolumn{2}{c}{Segmentation} & \multicolumn{2}{c}{Depth} \\
		    & & &  \multicolumn{2}{c}{{(Higher Better $\uparrow$)}}
		     & \multicolumn{2}{c}{{(Lower Better $\downarrow$)}}\\
		    & & & mIoU & Pix Acc & Abs Err & Rel Err\\
		    \midrule
		    \multirow{7}{*}{SegNet} & STL & -\ - & 51.85 & 91.08 & 0.0136 & 22.68   \\
		    \cmidrule{2-7}
		     & \multirow{5}{*}{MTL} & Uniform & 50.73 & 90.76 & 0.0151 & 40.81\\
		     & & Uncert \cite{kendall2018multi} & 51.09 & 90.85 & 0.0143 & 27.66 \\
			 & & MGDA \cite{sener2018multi} & 51.69 & 90.99 & {\bf 0.0130} & {\bf 24.04} \\
			 & & GradNorm \cite{chen2017gradnorm} & 50.06 & 90.82 & 0.0143 & 28.61 \\
			 & & Ours & {\bf 52.18} & {\bf 91.24} & 0.0140 & 28.90 \\
		    \midrule
		    \multirow{8}{*}{MTAN} & STL & -\ - & 51.24 & 91.16 & 0.0137 & 24.80   \\
		    \cmidrule{2-7}
		     & \multirow{6}{*}{MTL} & Uniform & 52.56 & 91.33 & 0.0152 & 24.64 \\
		     & & DWA \cite{liu2019end} & 51.95 & 91.33 & 0.0141 & 30.03 \\
		     & & Uncert \cite{kendall2018multi} & 50.37 & 91.11 & 0.0142 & 31.78 \\
			 & & MGDA \cite{sener2018multi} & 52.32 & {\bf 91.59} & {\bf 0.0138} & 30.35 \\
			 & & GradNorm \cite{chen2017gradnorm} & 51.88 & 91.40 & 0.0148 & 31.43 \\
			 & & Ours & {\bf 52.71} & 91.54 & 0.0139 & {\bf 27.33} \\
		     \bottomrule
		\end{tabular}%
		}
		\vspace{-0.3cm}
		\caption{Testing results on Cityscapes.}
		\label{tab:cityscapes}
\end{table}%

\paragraph{Results on Cityscapes.}
Similar to NYU-V2, we use cross-entropy loss for semantic segmentation and l1-norm loss for depth estimation in Cityscapes as in \cite{liu2019end}.
We train all methods using Adam \cite{kingma2014adam} with the learning rate initialized at 1e-4 and halved at the 100-th epoch for 200 epochs in total. The learning rate of all task-specific adaptors is 0.1 and the batch size is 8. In our method, weights of task-specific losses (\ie $w$ in \Cref{eq:KDMT}) are set uniformly.
As a validation set for hyperparameter search ($\lambda$), we randomly pick 10\% of training data. After the best hyperparameters are chosen, we retrain with the full training set and report the results of two tasks on the validation set in \Cref{tab:cityscapes}. We search over the set $\lambda=\{1,2,3,4,5,6\}$ of $\lambda$ and the distillation loss' weight of Segmentation and depth are set to 2 and 6, respectively.

As shown in \Cref{tab:cityscapes}, in overall, MTL obtains worse performance than STL. It is clear that GradNorm obtains worse performance on semantic segmentation while it improves the performance on depth estimation. However, MGDA assigns much larger weight on depth estimation task and this enables the MTL model to achieve better performance on both tasks. Our method also achieves significant gains on both tasks. The results again demonstrate that our method is able to optimize MTL model in a more balanced way and to achieve better overall results.

\subsection{Ablation Study}

To better analyze the effect of the distillation loss, we conduct an ablation study on NYU-V2. We first evaluate the effect of applying distillation loss to more layer's features. On NYU-V2, we report the results of applying distillation loss to the last shared layer's feature only and the results of applying distillation loss to features of both the middle layer and the last layer. From the results presented in \Cref{tab:ablanyuv2}, adding more layers' features for computing distillation loss boosts the performance on NYU-V2 in general. However, results on SVHN \& Omniglot and Cityscapes indicate that using the last layers obtains the best performance. We argue that adding more layers can enhance the distillation loss and a more strict control on the parameters of the multi-task network. This is not necessary for those tasks that use a small network, \eg the network we used in SVHN \& Omniglot and would be useful for tasks using a large network, \eg the SegNet used in NYU-V2.

\begin{table}[t]
	\centering
	
    \resizebox{1.0\textwidth}{!}
	{
		\begin{tabular}{ccccccccccccc}


		    \toprule
		    & & & \multicolumn{2}{c}{Segmentation} & \multicolumn{2}{c}{Depth} & \multicolumn{5}{c}{Surface Normal} \\
		    \multicolumn{3}{c}{Method}&  \multicolumn{2}{c}{\multirow{2}{*}{(Higher Better $\uparrow$)}}
		     & \multicolumn{2}{c}{\multirow{2}{*}{(Lower Better $\downarrow$)}}
		    & \multicolumn{2}{c}{Angle Distance} & \multicolumn{3}{c}{Within $t^{\circ}$}\\
		    &  &  & & & & &  \multicolumn{2}{c}{(Lower Better $\downarrow$)} & \multicolumn{3}{c}{(Higher Better $\uparrow$)}\\
		    backbone & \#layers & adaptors  & mIoU & Pix Acc & Abs Err & Rel Err & Mean & Median & 11.25 & 22.5 & 30 \\
		    \midrule
		    \multirow{4}{*}{SegNet} & last & linear & 18.66 & 57.78 & 0.5813 & {\bf 0.2375} & 30.17 & 24.74 & 22.96 & 46.44 & 58.76\\
		     & mid + last & linear & 18.75 & 58.02 & {\bf 0.5780} & 0.2467 & {\bf 29.40} & {\bf 23.71} & {\bf 24.33} & {\bf 48.22} & {\bf 60.45}\\
		     & mid + last & non-linear & {\bf 19.00} & {\bf 58.12} & 0.5853 & 0.2398 & 29.74 & 24.25 & 23.06 & 47.16 & 59.53\\
		     & mid + last & \xmark & 17.11 & 54.60 & 0.6060 & 0.2586 & 30.32 & 24.74 & 22.66 & 46.36 & 58.72\\
		 	\midrule
		    \multirow{4}{*}{MTAN \cite{liu2019end}}  & last & linear & 18.52 & 56.81 & {\bf 0.5756} & 0.2489 & 31.13 & 24.93 & 23.52 & 46.29 &58.18 \\
		     & mid + last & linear & {\bf 20.75} & {\bf 57.90} & 0.5816 & {\bf 0.2445} & {\bf 29.97} & {\bf 23.96} & {\bf 24.24} & {\bf 47.78} & {\bf 59.78} \\
		     & mid + last & non-linear & 20.42 & 56.93 & 0.5975 & 0.2589 & 30.32 & 24.86 & 22.98 & 46.21 & 58.55 \\
		     & mid + last & \xmark & 19.30 & 53.85 & 0.6064 & 0.2567 & 31.43 & 27.61 & 17.61 & 40.86 & 54.63 \\
			\bottomrule
		\end{tabular}%
		}
		\vspace{-0.1cm}
		\caption{Ablation study on NYU-v2. Here, `\#layer' means which layers are selected for computing distillation loss. `adaptors' indicates which kind of adaptors is used and `\xmark' means no adaptors are used.}
		\label{tab:ablanyuv2}
\end{table}%

\paragraph{Analysis on Adaptors.} We evaluate our method using different types of adaptors. As mentioned in \Cref{sec:method}, we use a linear layer that consists of a $1\times1\times C\times C$ convolution for each adaptor and denote as `linear' in \Cref{tab:ablanyuv2}. We also evaluate our method without any adaptors (\ie indicated as `\xmark') and with `non-linear 'adaptors (\ie each adaptors consists of a $1\times 1 \times C \times 2C$ convolution, a Relu activation layer and a $1\times 1 \times 2C \times C$ convolution). From the results shown in \Cref{tab:ablanyuv2}, it is clear to see that the adaptors help to align the features of multi-task and single-task as each single-task model produce different features. Compared with our method using `non-linear' adaptors, which have much larger capacity of mapping features, the results indicate that using `linear' adaptors is sufficient as we discuss in \Cref{sec:method}.

\begin{figure}[t]
	\vspace{-0.2cm}
	\RawFloats
	\centering
	\begin{minipage}[t]{0.44\textwidth}
	\centering
	\includegraphics[width=0.99\linewidth]{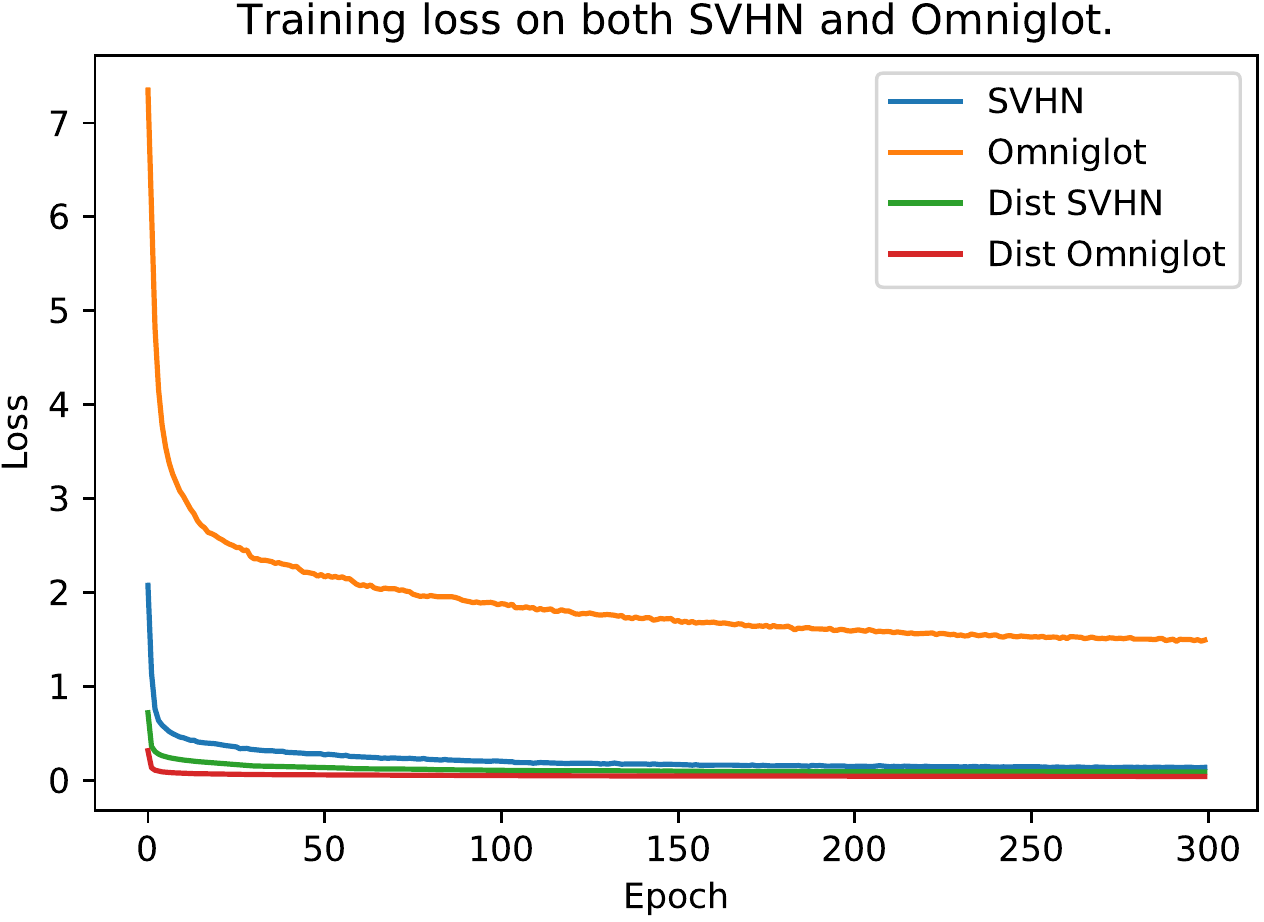}
	\vspace{-0.6cm}
	\caption{Task-specific loss and distillation loss on training set of SVHN \& Omniglot. Best view in color.}
	\label{fig:svhn}
	\end{minipage}
	\hspace{0.1cm}
	\begin{minipage}[t]{0.46\textwidth}
	\centering
	\includegraphics[width=0.99\linewidth]{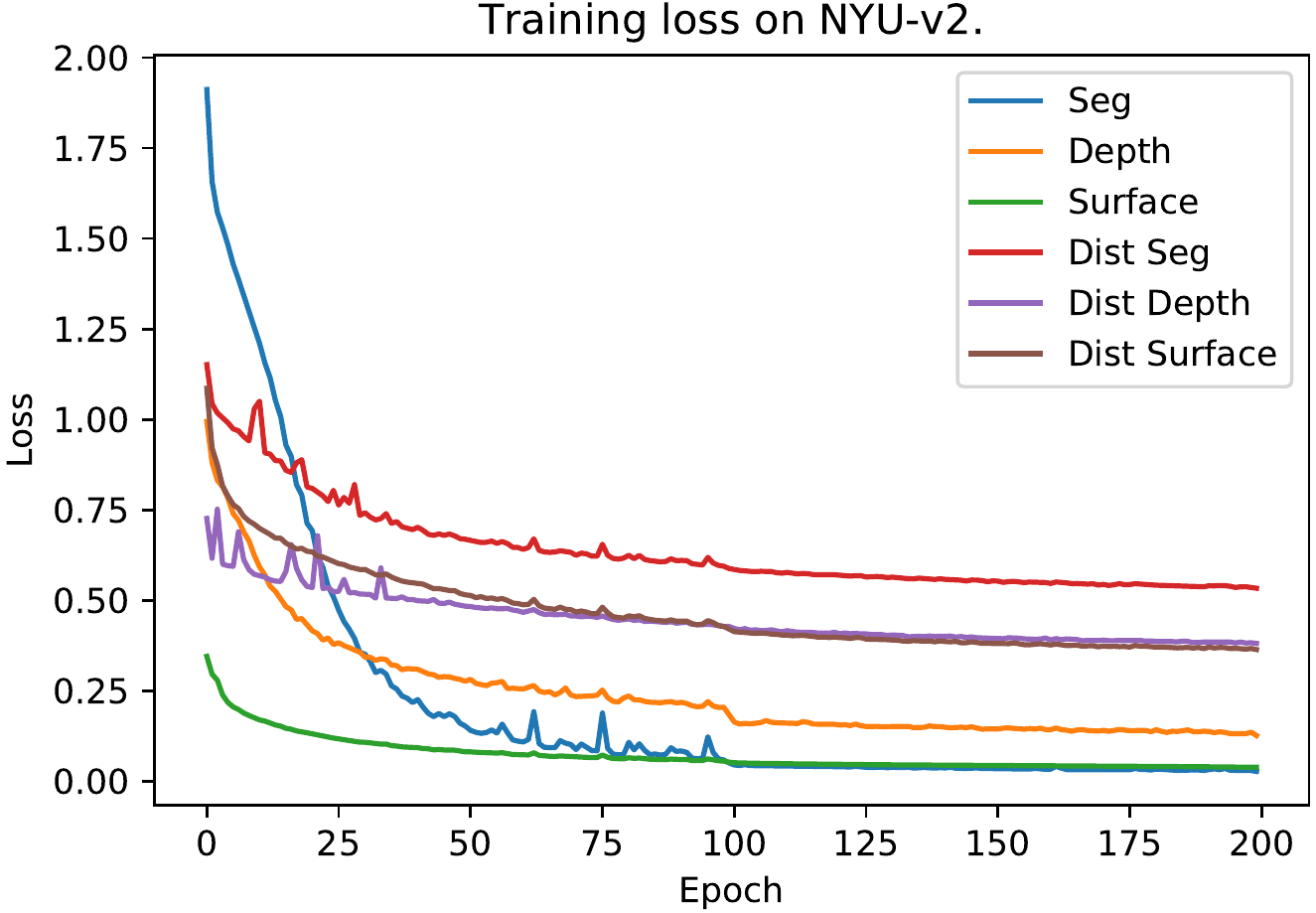}
	\vspace{-0.6cm}
	\caption{Task-specific loss and distillation loss on training set of NYU-V2. Best view in color.}
	\label{fig:nyu} 
	\end{minipage}
	\vspace{-0.4cm}
\end{figure}

\paragraph{Further Analysis.}
We also plot the task-specific loss and distillation loss on the training set of two benchmarks for analyzing our method. In both \Cref{fig:svhn} and \Cref{fig:nyu}, it is clear that distillation loss is more balanced than task-specific loss. More specifically, the task-specific loss of SVHN and Omniglot converge at around 0.137 and 1.492, respectively. In contrast, the distillation loss of SVHN and Omniglot converge at around 0.089 and 0.041, respectively. On NYU-V2, the task-specific loss of semantic segmentation, depth estimation, and surface normal estimation end up at 0.027, 0.127, and 0.039 while the distillation loss ends up at around 0.534, 0.381, and 0.364. These results again verify that our method can optimize the MTL method in a more balanced way. 



\section{Conclusion}
\label{sec:conc}

In this work, we proposed a knowledge distillation based multi-task method that learns to produce the same features with the single-task networks to address the unbalanced multi-task learning problem with the hypothesis that the solution of the multi-task network should be close to the single task ones' and lie in the intersection of the single solutions. We demonstrated that our method achieves significant performance gains over the state-of-the-art methods, on challenging benchmarks for image classification and scene understanding (semantic segmentation, depth estimation, and surface normal estimation). As future work, we plan to extend our method to multi-task network architecture searching.

\newpage

\bibliographystyle{splncs04}
\bibliography{refs}

\end{document}